\documentclass[10pt,twocolumn,letterpaper]{article}

\usepackage[preprint]{iccv}      
\usepackage{multirow}
\usepackage{amsmath}

\usepackage{amsmath,amsfonts, bm}

\def\1{\bm{1}}

\def\vx{{\bm{x}}}

\DeclareMathAlphabet{\mathsfit}{\encodingdefault}{\sfdefault}{m}{sl}
\SetMathAlphabet{\mathsfit}{bold}{\encodingdefault}{\sfdefault}{bx}{n}

\definecolor{iccvblue}{rgb}{0.21,0.49,0.74}
\usepackage[pagebackref,breaklinks,colorlinks,allcolors=iccvblue]{hyperref}

\def\paperID{*****} 
\def\confName{ICCV}
\def\confYear{2025}

\title{UniTEX: Universal High Fidelity Generative Texturing for 3D Shapes}
\vspace{-16pt}
\author{%
Yixun Liang$^{1,2}$\footnotemark[1] \quad 
Kunming Luo$^{1,2}$\footnotemark[1] \quad
Xiao Chen$^{2}$\footnotemark[1] \quad \\
Rui Chen$^{1}$ \quad
Hongyu Yan$^{1}$ \quad
Weiyu Li$^{1,2}$ \quad 
Jiarui Liu$^{1,2}$ \quad
Ping Tan$^{1,2}$\footnotemark[2]
\\ \vspace{-10pt}\\
$^{1}$HKUST \qquad 
$^{2}$Light Illusion \qquad 
}
\begin{document}


\twocolumn[{%
\renewcommand\twocolumn[1][]{#1}%
\maketitle
\begin{center}
    \centering
    \captionsetup{type=figure}
    \includegraphics[width=1.0\linewidth]{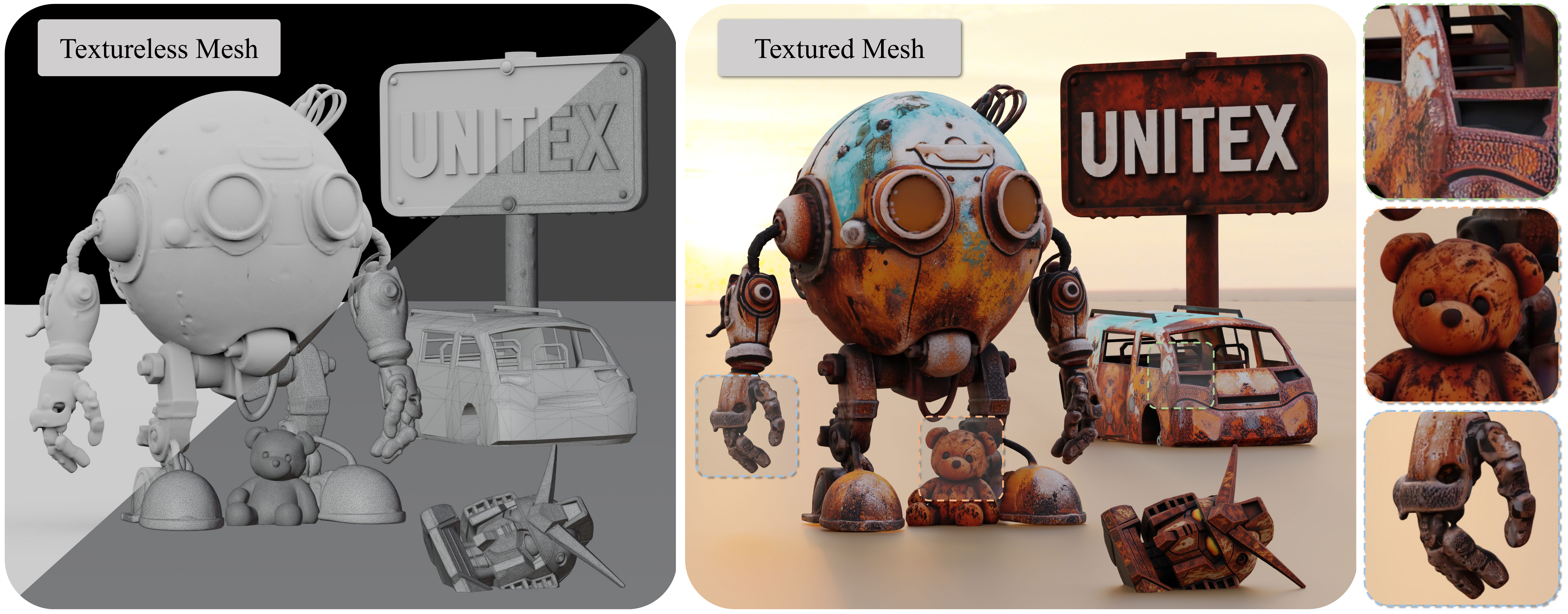}
    \vspace{-0.1in}
    \captionof{figure}{UniTEX generates \emph{high-quality} and \emph{complete} textures for both artist-created low-polygon mesh (the van shell) and generative high-polygon meshes (robots, toy bear, bust and roadsign).}
    \label{fig:teaser}
\end{center}%
}]

\renewcommand{\thefootnote}{\fnsymbol{footnote}} 
\footnotetext[1]{Equal contribution. Work done during internship at Light Illusion.} 
\footnotetext[2]{Corresponding authors.}

\begin{abstract}
We present UniTEX, a novel two-stage 3D texture generation framework to create high-quality, consistent textures for 3D assets.
Existing approaches predominantly rely on UV-based inpainting to refine textures after reprojecting the generated multi-view images onto the 3D shapes, which introduces challenges related to topological ambiguity. To address this, we propose to bypass the limitations of UV mapping by operating directly in a unified 3D functional space. Specifically, we first propose that lifts texture generation into 3D space via Texture Functions (TFs)—a continuous, volumetric representation that maps any 3D point to a texture value based solely on surface proximity, independent of mesh topology. Then, we propose to predict these TFs directly from images and geometry inputs using a transformer-based Large Texturing Model (LTM). To further enhance texture quality and leverage powerful 2D priors, we develop an advanced LoRA-based strategy for efficiently adapting large-scale Diffusion Transformers (DiTs) for high-quality multi-view texture synthesis as our first stage. Extensive experiments demonstrate that UniTEX achieves superior visual quality and texture integrity compared to existing approaches, offering a generalizable and scalable solution for automated 3D texture generation. Code will available in: \url{https://github.com/YixunLiang/UniTEX}
\end{abstract}    
\section{Introduction}
\label{sec:intro}

High-quality texture generation plays a critical role in 3D asset creation, as it directly impacts both the visual realism and semantic fidelity of rendered models. In modern applications such as gaming, virtual reality, and digital content creation, photorealistic textures are essential for delivering immersive and visually engaging experiences. However, producing detailed and aesthetically refined textures typically requires significant manual effort and domain expertise, making the process time-consuming and resource-intensive.

\begin{figure}[t]
    \includegraphics[width=1\linewidth]{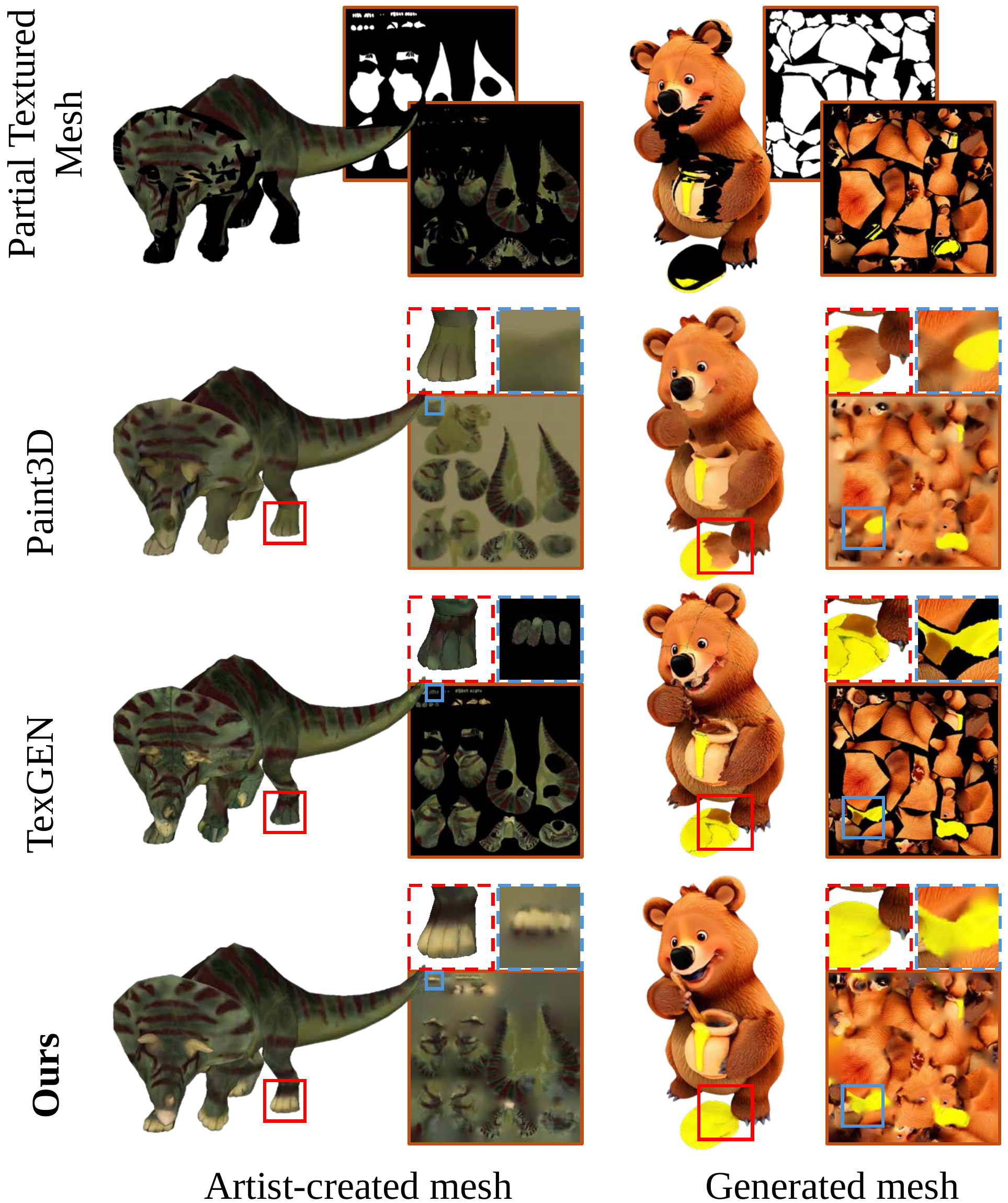}
    \centering
    \vspace{-0.7cm}
    \caption{UV-based texturing models perform well on in-domain, artist-created meshes (first column), but struggle with out-of-distribution, generated meshes (second column). We take Paint3D~\cite{zeng2024paint3d} and TexGEN~\cite{yu2024texgen} as representative examples: while effective on large, continuous regions, they fail to handle small, fragmented areas due to training biases toward clean, large-region UV layouts. In contrast, our method operates outside the UV space, enabling better generalization across diverse mesh types. Additional comparisons are provided in Sec.~\ref{sec:refinement_stage_comp}.} 
    \vspace{-0.75cm}
    \label{fig:uv_drawback}
\end{figure}

Recently, diffusion models trained on large-scale datasets~\cite{rombach2022high,esser2024sd3} have brought about a revolutionary transformation in the field of image and video generation. Inspired by this, there is an expectation that 3D texture generation can also be greatly simplified, enabling direct generation from text descriptions or a single reference image. Prior works~\cite{cheng2024mvpaint,zeng2024paint3d,bensadoun2024metatexgen,liu2024syncmvd} have primarily addressed this by adapting 2D generative models for multi-view texture generation and then projecting these multi-view textures onto the 3D mesh. While this approach leverages strong 2D priors and achieves some promising results, the generated textures often suffer from issues of self-occlusion and multi-view inconsistencies, leading to textures on the 3D surface that are incomplete or fragmented. To mitigate this, some early attempts~\cite{yu2024texgen,zeng2024paint3d,bensadoun2024metatexgen} introduce a second stage of UV-based inpainting process to merge the partial views into a coherent texture map to handle those incomplete or inconsistent areas after multi-view projection. Although these methods provide completed texture, in practice, UV-based inpainting often struggles to handle meshes created from generative AI pipelines~\cite{li2025triposg,li2024craftsman,zhao2025hunyuan3d} as shown in Fig.~\ref{fig:uv_drawback}. This practical observation highlights the generalization bottlenecks of UV-based inpainting for real-world 3D texture generation.

This challenge stems from a fundamental limitation of UV mapping, what we refer to as \textbf{topological ambiguity}. Specifically, a single 3D mesh can correspond to multiple valid UV layouts that are not uniquely determined by the geometry but are also highly sensitive to vertice/face distributions and UV unwarping algorithms. This ambiguity poses a critical challenge for UV-based inpainting models because there is no guarantee that the UV parameterization used during inference will match that of the training stage. In practice, mesh topology varies significantly due to differences in modeling conventions across designers. Consequently, the UV parameterization becomes inherently inconsistent. This problem is especially pronounced when texturing meshes produced by generative pipelines~\cite{li2025triposg,li2024craftsman,zhao2025hunyuan3d}, which are extracted using Marching Cubes~\cite{lorensen1998marching}. The resulting domain gap severely limits generalization during the inference of current UV-based inpainting models.
\label{sec:intro}

To address this challenge, we propose a novel approach to represent textures in a unified 3D functional space so as to bypass the limitations of refinement operations in UV space. Specifically, we introduce \textbf{Texture Functions (TFs)}—a continuous representation that maps any 3D spatial point to a texture value. For each query point in space, we determine the closest surface point on the mesh using a method analogous to computing Unsigned Distance Functions (UDF) and then retrieve the corresponding texture value from that location. While the texture is sampled based on surface proximity, TFs are defined throughout the entire 3D volume, enabling a volumetric texture representation. Unlike UV maps, which rely on mesh-specific face distribution and suffer from topological ambiguity, TFs are invariant to mesh face distributions and depend solely on surface position, which bypasses the problem we mentioned above. Also, this formulation allows texture to be treated as a smooth, continuous field over 3D space—analogous in spirit to how SDFs or UDFs represent geometry, but instead modeling appearance only on the surface like previous methods~\cite{oechsle2019texture,jiang2025dimer}. This brings a more complete training for texture generation, as detailed in Sec.~\ref{sec:tf_ablation}.

With this formulation, we frame texture inpainting/prediction as a native 3D regression task like~\cite{liu2023one12345++,liu2024meshformer}, where the model takes image and geometry inputs to directly predict the corresponding 3D texture functions. To realize this, we introduce a transformer-based architecture—\textbf{Large Texturing Model (LTM)}—which is detailed in Sec.~\ref{sec:ltm} to perform this task. By eliminating reliance on UV layouts, our approach reduces the domain gap between training data and real-world application, providing a generalizable and scalable second stage in the 3D texturing pipeline.

To further enhance the overall quality of the texture generation pipeline, we propose an advanced LoRA-based training strategy to efficiently adapt large-scale diffusion transformers (DiTs) for multiview synthesis conditioned on geometry and reference images. Since the first-stage texture generation heavily relies on the capabilities of 2D diffusion models—and large-scale DiT models currently dominate 2D generative modeling—our approach is designed to efficiently leverage these strengths. It facilitates scalable adaptation of powerful foundation models such as FLUX and SD3~\cite{esser2024sd3}, thereby improving generative texture quality and offering transferable insights for other vision tasks.

We evaluate our method through extensive experiments on multiple settings, demonstrating its superiority over existing approaches in terms of both visual quality and texture integrity. Overall, our contributions are summarized as:

\begin{itemize}
\item We propose \emph{Texture Functions (TFs)}, a continuous 3D texture representation that bypasses UV mapping and models texture as a completed spatial field.

\item We design a novel \emph{Large Texturing Model (LTM)} based on transformer architecture to predict TFs directly from images and geometry inputs.

\item We develop a LoRA-based strategy to efficiently adapt large diffusion transformers (DiTs) for downstream tasks, enabling high-quality multiview synthesis for texture generation using large-scale Diffusion Transformers.
\end{itemize}

\section{Related work}
\begin{figure*}[t]
\includegraphics[width=1\linewidth]{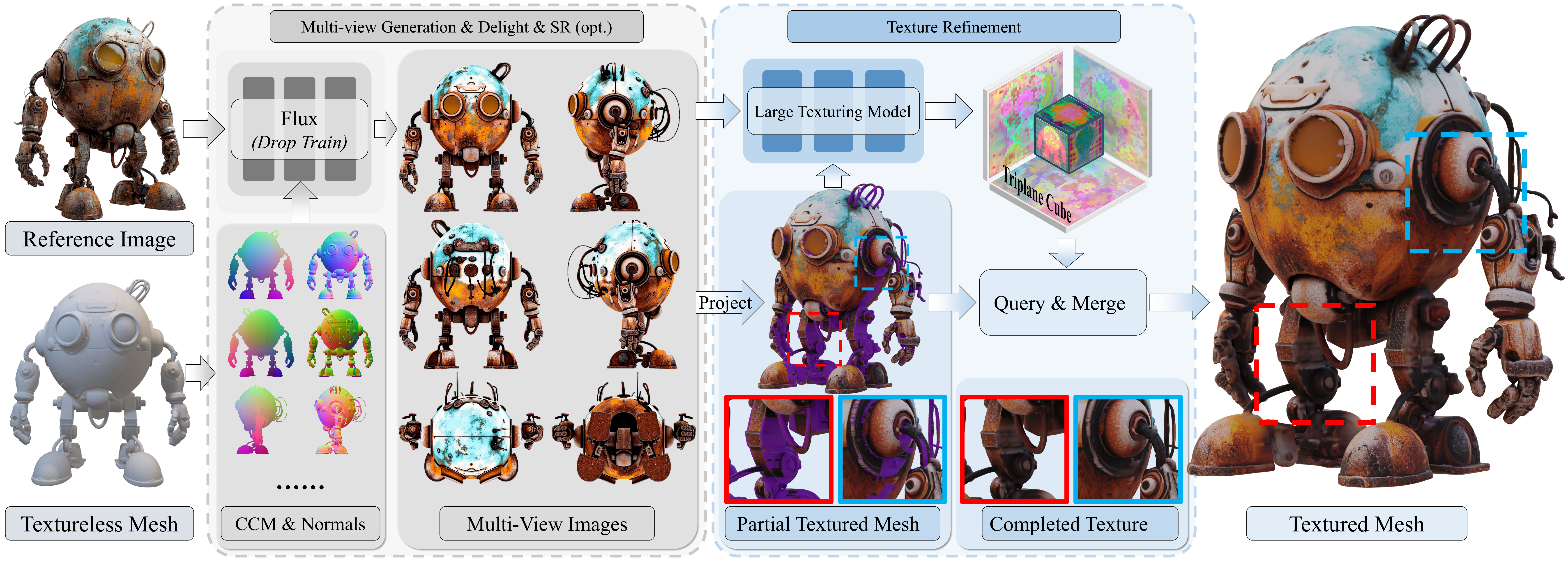}
    \centering
    \vspace{-0.5cm}
    \caption{Overall pipeline of UniTEX. Given a textureless geometry and reference image, UniTEX first generates a high-fidelity multi-view image through 3 steps (RGB generation, delighting, and super-resolution (SR)) using finetuned DiTs (detailed in Sec.~\ref{sec:imporve_training}). The texture will be reprojected to a partial textured mesh and sent to the Large Texturing Model (Detailed in Sec.~\ref{sec:ltm}) with generated images to predict the corresponding complete texture functions (Detailed in Sec.~\ref{sec:training_obj}). The final texture is then synthesized by blending the predicted texture functions with the partial textured geometry.}
        \vspace{-0.5cm}
    \label{fig:overall_pipeline}
\end{figure*}

\subsection{3D Texturing using Diffusion Models}
Most existing texturing methods leverage 2D priors from text-to-image (T2I) diffusion models~\cite{rombach2022high} to generate textures from text or images. Methods like Text2Tex~\cite{chen2023text2tex} and TEXture~\cite{richardson2023texture} iteratively paint meshes from multiple views but often lack multi-view consistency. Another line of work uses optimization-based approaches with Score Distillation Sampling (SDS) to directly refine texture maps, as seen in Fantasia3D~\cite{chen2023fantasia3d}, Magic3D~\cite{lin2023magic3d}, and DreamMat~\cite{zhang2024dreammat}. FlashTex~\cite{deng2024flashtex} improves light disentanglement by introducing a light-conditioned diffusion model in a two-stage pipeline. However, SDS-based methods typically rely on single-view diffusion models, suffer from view inconsistency (e.g., Janus problem), and require expensive iterative optimization, limiting their scalability.

Subsequent works aim to improve multiview consistency and texture quality by fine-tuning diffusion models~\cite{zhang2024clay,zhao2025hunyuan3d,zeng2024paint3d,cheng2024mvpaint}, introducing advanced sampling strategies~\cite{zhang2024texpainter,cao2023texfusion,liu2024syncmvd}, or designing 2D diffusion models to adopt UV refinement~\cite{bensadoun2024metatexgen,zeng2024paint3d,yu2024texgen}. TexFusion~\cite{cao2023texfusion} uses a denoising sampler that fuses images across views in UV space at each step. Paint3D~\cite{zeng2024paint3d} refines textures through a UV inpainting model. Meta 3D TextureGen~\cite{bensadoun2024metatexgen} leverages a geometry-aware T2I model for consistent multi-view synthesis, and Hunyuan3D 2.0~\cite{zhao2025hunyuan3d} proposes a model to generate multi-view textures simultaneously. 
Although these UV-based methods can leverage the powerful priors of 2D diffusion models, the UV representation inherently suffers from topological ambiguity, which limits the generalization capabilities of these methods. In this paper, we propose Texture Functions (TFs), a continuous 3D texture representation, aiming to bypass the limitations of UV unwrapping and achieve complete and high-quality texturing.

\subsection{3D Native Texturing}
Another line of work trains generative models directly on 3D data with ground-truth textures~\cite{oechsle2019texture,yu2024texgen,yu2023texture,Liu2024TexOct,xiong2024texgaussian}. Texture Field~\cite{oechsle2019texture} learns implicit color fields on 3D surfaces, while Texturify~\cite{siddiqui2022texturify} uses face convolutions and adversarial rendering losses for per-face texture prediction. Recent diffusion-based methods such as PointUV~\cite{yu2023texture}, TexGEN~\cite{yu2024texgen}, and TexOct~\cite{Liu2024TexOct} generate point cloud colors mapped to UV textures. TexGaussian~\cite{xiong2024texgaussian} introduces an octree-based 3D Gaussian representation trained to predict textures from text and geometry and then bake them to the original mesh. Although these native 3D approaches offer better completion compared with 2D diffusion-based methods, they are limited by the scarcity of 3D data and limited generalizability. In this paper, we reposition native 3D texturing as a refinement module in a second-stage refinement and get more generative conditions (multi-view images) from 2D foundation models like~\cite{li2023instant3d}. By integrating the strong generative capabilities of 2D diffusion models, our method achieves more robust performance under limited data and produces higher-quality results.

\section{Methods}
\subsection{Overview}
In this section, we present a detailed design of our texturing pipeline, which is a \textbf{two-stage} pipeline that fully considers the combination of the 2D diffusion models for Multiview generation and 3D texturing methods for texture completion. As shown in Fig.~\ref{fig:overall_pipeline}, given a single input image and a textureless 3D mesh, our system first fine-tunes two large-scale diffusion transformers (Flux 1~\footnote{https://huggingface.co/black-forest-labs/FLUX.1-dev}) using an efficient LoRA-based training strategy (Sec.~\ref{sec:imporve_training}) to generate six orthographic, illumination-free views conditioned on the mesh's normals and canonical coordinate map (CCM)~\cite{li2023sweetdreamer,wang2024crm}, which is a special type of rendered image where each pixel encodes the 3D coordinate of the surface point it corresponds to). These generated images can optionally be used for super-resolution~\cite{dong2024tsd}. After reprojection and blending, the synthesized views, along with the partially textured geometry, are fed into our Large Texturing Model (LTM) (Sec.~\ref{sec:ltm}), which predicts the corresponding complete texture functions(Sec.~\ref{sec:training_obj}). The final texture is then synthesized by blending the predicted texture functions with the initial partial textured geometry.
\begin{figure}[t]
\includegraphics[width=1\linewidth]{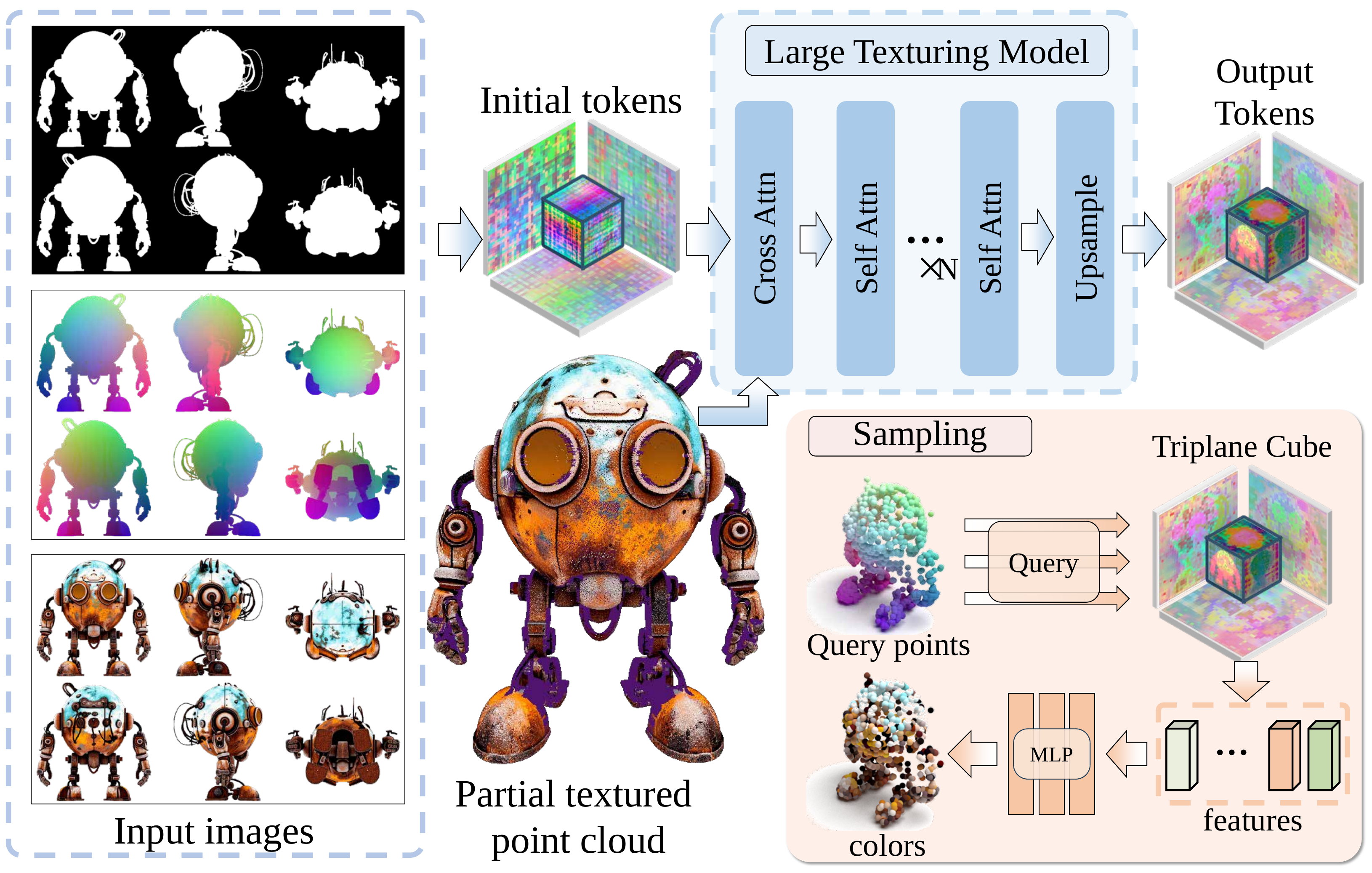}
    \centering
    \vspace{-0.75cm}
    \caption{Pipeline of the Large Texturing Model. Given a partially textured geometry and six input views, we first unify them into a shared triplane-cube token representation. A transformer-based architecture then processes these tokens to extract geometry-aware features, which are subsequently decoded into colors using a lightweight MLP.}
    \label{fig:LTM_pipeline}
\vspace{-0.5cm}
\end{figure}
\subsection{Efficient DiT Tuning for MV Generation}
\label{sec:imporve_training}
In this section, we present the details of the first stage in Fig~\ref{fig:overall_pipeline}. We fine-tune two large-scale diffusion transformers using in-context learning to adapt the model to our specific task -- 3D texturing. Specifically, the first Flux takes reference images, normal maps, and canonical coordinate maps (CCMs) of textureless mesh as input and generates corresponding shaded images. The second Flux is used to delight and generate diffuse color for final texturing. The generated diffuse color images can be used with super-resolution~\cite{dong2024tsd} for better results.

However, effectively adapting 2D diffusion models for 3D texturing presents unique challenges. Unlike traditional 2D generation, our task requires significantly more conditioning signals. For instance, generating six-view images at 512×512 resolution demands a reference image and corresponding geometric information (e.g., normals and CCMs) for all six views. Moreover, since diffusion transformers (DiTs) rely heavily on in-context learning~\cite{zhang2025easycontrol, tan2024ominicontrol}—where all inputs and conditions are jointly encoded as tokens—the resulting volume of token inputs substantially increases training cost and slows convergence.

Recent studies, such as MVDiffusion++\cite{tang2024mvdiffusion++} and Long-LoRA\cite{chen2023longlora}, suggest that it is not strictly necessary to preserve the full forward computation pattern during both training and inference when fine-tuning large-scale foundation models. Specifically, Long-LoRA proposes truncating attention windows during training while restoring full attention at inference time to save resources and achieve long context finetuning. Similarly, MVDiffusion++ demonstrates that training on fewer views and inferring more views at test time can still yield strong performance. This invites a rethinking of why such behaviors can work and what fine-tuning is truly optimizing for. Using texture generation as an example, the image quality does not need to be trained during finetuning, the pattern we want the model to learn is "multi-view consistency and correspond to conditions". Learning such a pattern may not require simultaneous access to all input tokens. Building on this insight, we propose a drop training strategy: during each training step, only a subset of all tokens is retained, and the diffusion transformer is conditioned and generated solely on these selected tokens rather than the full input tokens. This approach reduces the dependency on complete image tokens, allowing the model to learn from partial information while maintaining task-relevant needs. As shown in Sec.~\ref{sec:diffusion_training_strategy}, This method achieves comparable generation quality to full-input fine-tuning at the same iterations while significantly accelerating training and reducing the computational cost.

\subsection{Large Texturing Model}
\label{sec:ltm}
Previous two-stage texturing methods primarily relied on UV-based inpainting, which often suffers from topological ambiguities and leads to suboptimal texture quality. In our work, we propose \textit{Large Texturing Model (LTM)} to regress the texture in 3D functional space to bypass the topology ambiguity and serve as the second stage in Fig.~\ref{fig:overall_pipeline}.

The visualization of our \textit{Large Texturing Model (LTM)} is shown in Fig.~\ref{fig:LTM_pipeline}.
Based on the generated images from Sec.~\ref{sec:imporve_training} with textureless/ incompleted texture geometry, we introduce the \textit{Large Texturing Model (LTM)}, which architecture design is detailed in Sec.~\ref{lab:ach_design}), which is designed to regress the \textit{Texture Functions} (Detailed in Sec.~\ref{sec:training_obj}) for the second stage of our texturing pipeline. 

\subsubsection{Achitechture Design}
\label{lab:ach_design}
We first initialize our \textit{triplane-cube} with a set of learnable tokens, which contain a high-resolution triplane $\mathcal{T} \in \mathbb{R}^{3\times 32\times 32}$. and a low-resolution cube $\mathcal{C} \in \mathbb{R}^{8\times 8\times 8}$, which have been extensively demonstrated to offer a superior representation compared to triplanes~\cite{reiser2023merf,guo2025hyper3d}. Then, we encode 6 orth-view images, CCM, and alpha maps to a triplane like CRM~\cite{wang2024crm} and add them together to initial triplane tokens. After that, we encode geometry information to such representation following Shape2VecSet~\cite{zhang20233dshape2vecset}. Specifically, we sample the colored point cloud in the partial textured geometry and use the \textit{triplane-cube} tokens to query them through cross-attention.

After integrating the information of 2D images and 3D geometry, we use a transformer architecture with self-attention to process these flatted features of \textit{triplane-cube}. After processing such representation. The final output is first reshaped into triplane and cube representations, then upsampled using 2D and 3D deconvolution layers respectively.

During sampling, We implement an MLP, denoted as $\text{MLP}_{\theta}$ to predict RGB queried from the triplane-cube features denoted as $\mathcal{TC}$. Given a query 3D point $\vx$, we predict the correspondence color as:
\begin{equation}
    \hat{\mathbf{c}}(\vx, \mathcal{TC})= \text{MLP}_{\theta}(\mathit{gridsample}(\vx, \mathcal{TC})),
\end{equation}
where $\hat{\mathbf{c}} \in \mathbb{R}^3$ is the predicted RGB color for point $\vx$.

\begin{figure}[t]
    \includegraphics[width=1\linewidth]{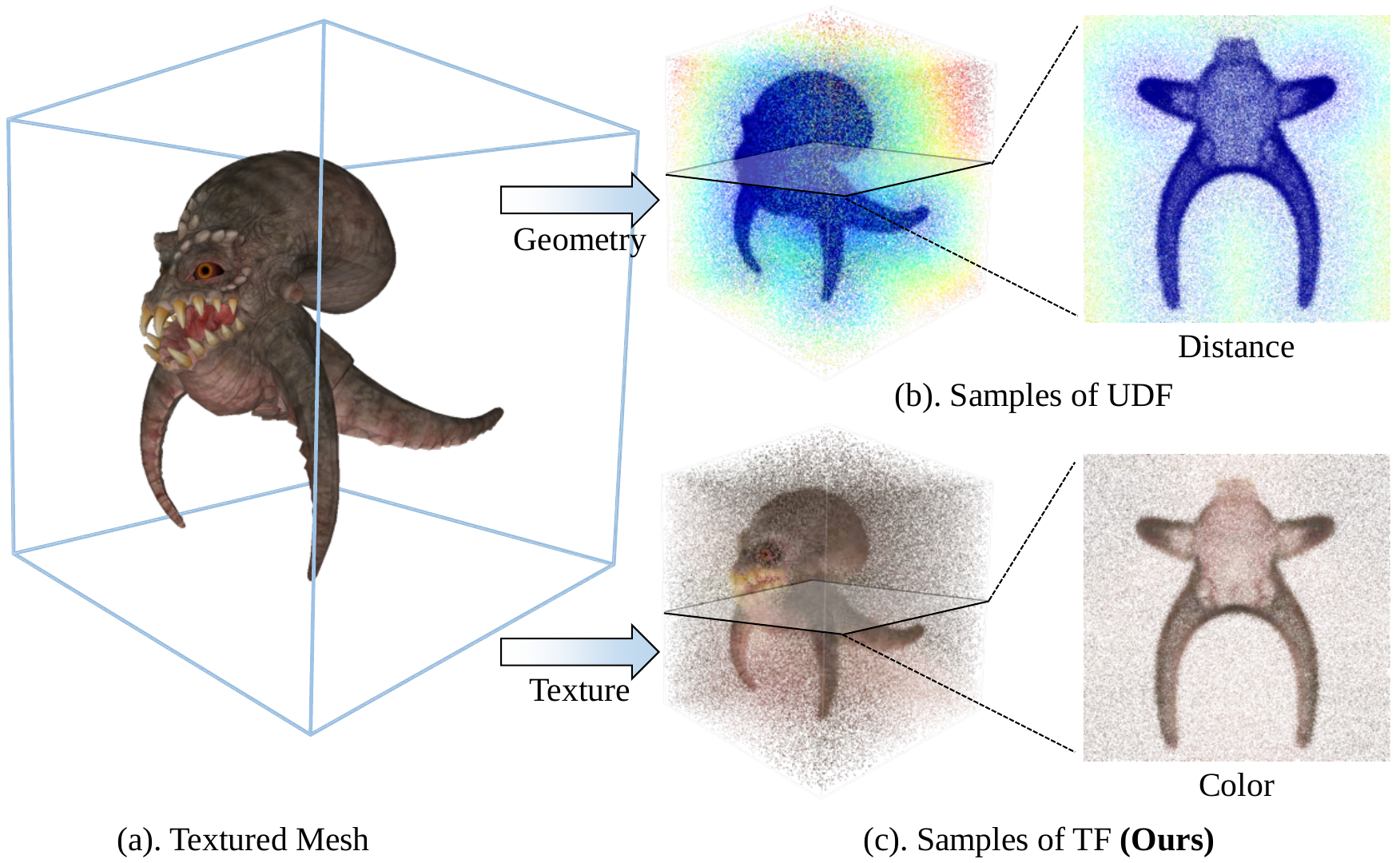}
    \centering
    \vspace{-0.7cm}
    \caption{Visualized example of the \textit{Texture Functions} (TF) for represent the texture for the whole 3D space. (a) A textured mesh. (b) Unsigned Distance Function (UDF) samples representing 3D geometry. (c) Inspired by UDF, we define texture as a continuous function over 3D space (details in Sec.~\ref{sec:training_obj}), enabling volumetric texture representation.}
    \vspace{-0.7cm}
    \label{fig:explaination_texfuncs}
\end{figure}
\subsubsection{Training Objective -- Texture Functions:} 
\label{sec:training_obj}
Unlike native 3D geometry generation or reconstruction, which benefits from well-defined Signed or Unsigned Distance Fields (SDF/UDF) in 3D functional space as continuous and complete supervision signals~\cite{li2024craftsman, li2025triposg, zhao2025hunyuan3d}, texture is traditionally defined only on the surface of 3D objects. Prior works~\cite{oechsle2019texture, siddiqui2022texturify, hong2023lrm, li2023instant3d,jiang2025dimer} typically rely on surface or volume rendering to supervise the texture representation via 2D projections. However, this form of supervision is inherently sparse and limited in coverage compared to the dense volumetric supervision available in geometry tasks. As shown in geometry generation, complete 3D supervision—such as that provided by SDFs/UDFs—consistently outperforms sparse alternatives like LRM-based methods.
\begin{figure*}[t]
\centering
  \includegraphics[width=1.0\textwidth]{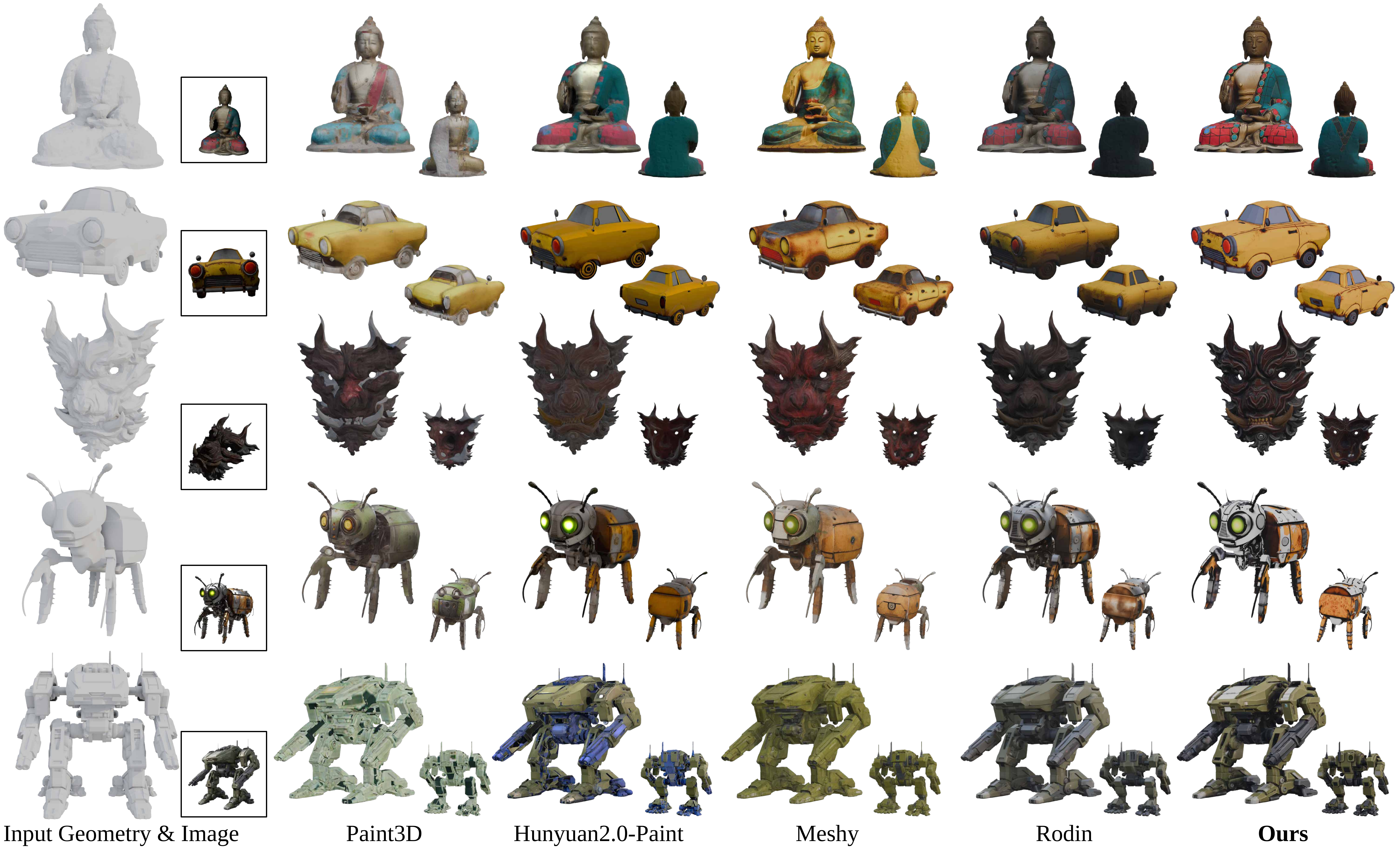}
  \vspace{-0.7cm}
  \caption{Qualitative comparison of our method against state-of-the-art (Paint3D~\cite{zeng2024paint3d}, Hunyuan2.0-Paint~\cite{zhao2025hunyuan3d}) and commercial proprietary (Rodin, Meshy) texturing approaches. Our method consistently achieves superior texture quality and generalization across diverse mesh sources. (best viewed by zoom in)
}
  \label{fig:main_comparision}
\end{figure*}
Motivated by this insight, we propose to extend texture from a surface-restricted signal to a continuous volumetric function defined throughout the 3D space. This allows us to supervise the texture model using densely sampled points across the volume, providing richer and more complete training signals. Formally, we define a Texture Function as a mapping over 3D coordinates $\vx \in \mathbb{R}^3$, where each texture value is obtained by orthogonally projecting $\vx$ onto the closest point on the mesh surface $\Omega$ and querying its corresponding color. This definition aligns naturally with volumetric geometry representations and enables unified learning across the entire 3D domain. For better understanding, Fig.~\ref{fig:explaination_texfuncs} provides a visual comparison between our proposed texture function and the traditional unsigned distance function (UDF). 

With such Texture functions, the training objectives of LTM is defined as:
\begin{equation}
    \mathcal{L}_{\text{texture}} = \mathbb{E}_{\mathbf{x} \sim \Omega} \left[ \|\hat{\mathbf{c}}(\vx, \mathcal{TC}) - \mathbf{c}(\vx)^*\|^2 \right] + \lambda\mathcal{L}_{tv}(\mathcal{T}),
\end{equation}
where $\vx$ is the set of surface points, $\hat{\mathbf{c}}(\mathbf{x})$ is the predicted color, and $\mathbf{c}(\vx)$ is the ground truth color derive from texture functions. $\mathcal{L}_{tv}$ denote as total-variance loss to regulize the variance of grid-based 3D representation. The $\lambda$ denotes the weight, which we set as $0.0005$ in our experiments. To be noticed, we use truncated texture functions like truncated SDF that are used for training in~\cite{zhao2025hunyuan3d,li2025triposg,chen2024dora}. We set the threshold as $0.025$. For those points that are out of the truncated value, we use background color as the ground truth for training.

In addition to providing surface point supervision similar to previous methods, our approach further extends supervision to non-geometric regions in 3D space. Specifically, the colored regions are expanded into a thin shell (after truncation) surrounding the geometry. This brings notable advantages: during color prediction, the model is implicitly encouraged to build a volumetric understanding of the 3D object. The introduction of the thin shell alleviates the need for highly precise mesh modeling, as the model can still correctly query colors without relying on exact geometry. Moreover, a fully defined supervision signal is beneficial for learning a well-structured latent space and enhancing the model's generalizability. This leads to improved predicted texture quality, as demonstrated by our experiment results in Sec.~\ref{sec:tf_ablation}.
\section{Experiments}
\begin{table}[t]
\centering
\resizebox{\linewidth}{!}{%
\begin{tabular}{lcccccc}
\toprule
\multirow{2}{*}{Methods.} & \multicolumn{4}{c}{Artist-created Mesh} & \multicolumn{2}{c}{Generative Mesh} \\
\cmidrule(lr){2-5} \cmidrule(lr){6-7}
& CMMD$\downarrow$  & FID$_{\text{CLIP}}$$\downarrow$  & CLIP$_{\text{score}}$$\uparrow$  & LPIPS$\downarrow$  & CLIP$_{\text{score}}$$\uparrow$ & User-perf.$\uparrow$ \\
\midrule
Paint3D   & 1.196 & 20.52 & 0.840 & 0.107 & 0.737 & 6.82\% \\
TexPainter   & 1.632 & 27.13 & 0.804 & 0.112 &  0.773 & 1.36\% \\
TexGaussians   & 1.290 & 21.08 & 0.836 & 0.095 & 0.718 & 4.55\% \\
Hunyuan3D-Paint   & 0.909 & 20.38 & 0.824 &0.091& 0.805& 21.36\% \\
Ours   & \textbf{0.826} & \textbf{16.03} & \textbf{0.844} & \textbf{0.090} & \textbf{0.808} & \textbf{65.91\%}\\
\bottomrule
\end{tabular}%
}
\caption{Quantitative comparison between artist-created and generative mesh on different texturing methods.}
\vspace{-0.5cm}
\label{tab:texture_main_comparison}
\end{table}
\subsection{Experimental Setup}
\label{sec:tex_main_experiments}
\begin{figure*}[t]
\centering
\includegraphics[width=0.95\linewidth]{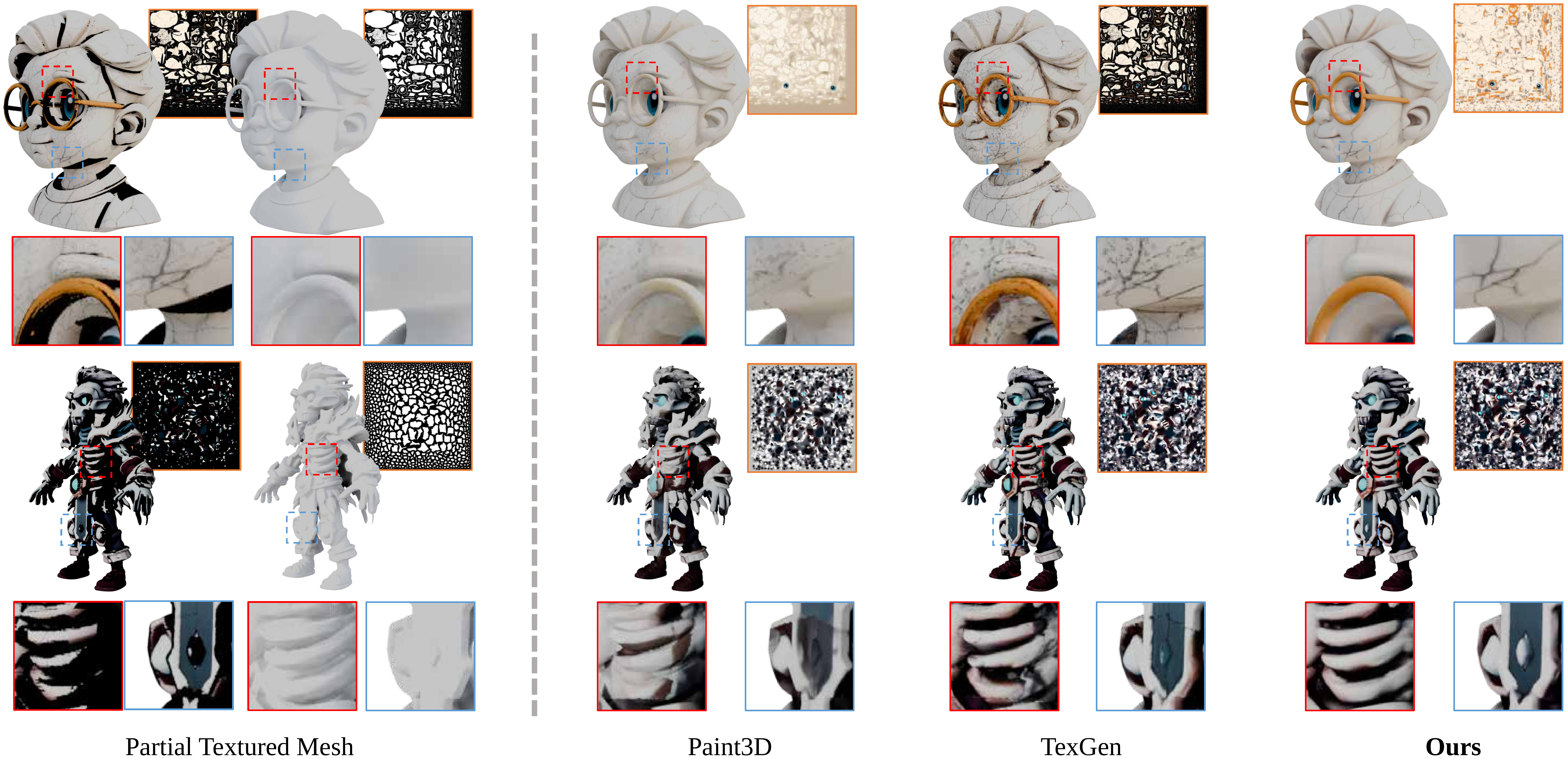}
\vspace{-0.25cm}
    \caption{Qualitative comparison of refinement stage across different methods. In the first case, automatically unwrapping makes \textbf{fragmented and noisy} UV layout on the face. UV-based methods such as Paint3D and TexGen struggle with these (blue boxs). In contrast, our method generates smooth and coherent textures that more respect to geometry (glasses in the first row and the ribcage and emblem in the second row).}
\label{fig:2nd_stage_comp}
\label{fig:refinement_stage_comp}
\end{figure*}
\paragraph{Baselines} In this section, we compare our overall pipeline with current open-source state-of-the-art (SoTA) methods, including Hunyuan3D-Paint~\cite{zhao2025hunyuan3d}, TexPainter~\cite{zhang2024texpainter}, TexGaussians~\cite{zhang2024texpainter}, and Paint3D~\cite{zeng2024paint3d} in quantitative comparison, in the qualitative comparison, we further compare our method with the closed-source method including Tripo, Hyper3D-Rodin~\footnote{The API was invoked from Tripo/Rodin platform in May 2025 and some of the meshes are downloaded from gallery.}. To be noticed, for the pipelines that can only receive text prompt as input, we use GPT-4o to caption the image and use it as input prompt, otherwise, use the reference image as input prompt. 

For the second stage comparison, we use the same partial textured model reprojected from six views and a reference image as input and use different methods to paint/complete this partial texture. We primarily compare our second stage with Paint3D-2nd stage~\cite{zeng2024paint3d}. Additionally, we observe that the settings of TexGen~\cite{yu2024texgen} can also be viewed as a second-stage model for UV-based inpainting, and thus, we include it in our second-stage comparisons as well. 

\begin{table}[!t]
\centering
\tiny
\resizebox{\linewidth}{!}{%
\begin{tabular}{c|ccccc}
\toprule
Methods.& PSNR\textsubscript{uv}$\uparrow$&PSNR\textsubscript{uv}$^{*}$$\uparrow$& PSNR$\uparrow$ & SSIM$\uparrow$ & LPIPS$\downarrow$\\
\midrule
 Paint3D-UVpaint & 18.61  &  15.22 & 23.90 & 0.971 & 0.042 \\
 TexGEN  &20.47 &  17.07  &  25.19 & 0.977 & 0.040\\
  \textbf{Ours}  & \textbf{23.01} &  \textbf{19.89}  & \textbf{30.45} & \textbf{0.989} & \textbf{0.023} \\
\bottomrule
\end{tabular}}
\caption{Refinement Stage Comparision, PSNR\textsubscript{uv}$^*$ indicates the PSNR calculated exclusively in the invisible regions, whereas PSNR\textsubscript{uv} is computed over the entire set of valid UV regions.}
\vspace{-0.5cm}
\label{tab:stage2comp}
\end{table}
\begin{figure*}[!t]
\centering
  \includegraphics[width=0.95\textwidth]{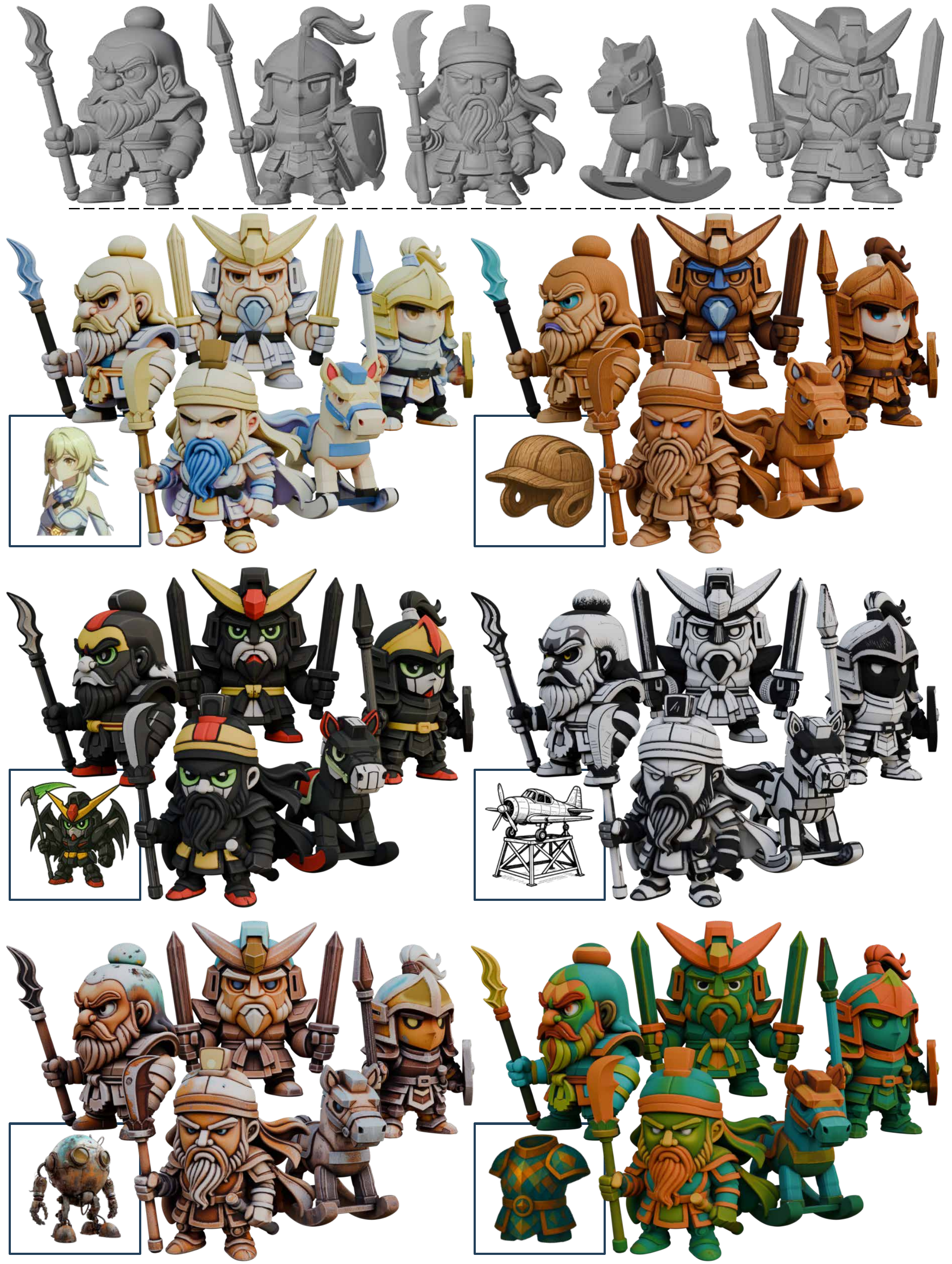}
  \vspace{-0.5cm}
  \caption{Qualitative results of stylized texturing. We evaluate our method with various input images and our method can robustly generate high-fidelity stylized textures.(Mesh generated by Hunyuan2.5)}
  \label{fig_only_style}
    \vspace{-0.75cm}
\end{figure*}
Unlike previous methods~\cite{zhao2025hunyuan3d,bensadoun2024metatexgen} contains several settings and some of them solely compare the texture generation ability in artist-created meshes. We set two benchmarks that consider both artist-created meshes and generative meshes. For artist-created meshes, we randomly select 78 meshes from objaverse~\cite{deitke2023objaverse} and the test list provided by MetaTexGEN~\cite{bensadoun2024metatexgen}. For generative meshes, we select 30 images and generate mesh using Craftsman~\cite{li2024craftsman} for texturing. This enables us to fully evaluate the texture generation ability in various real-world applications.
\paragraph{Evaluation Criteria} We adopt several widely used image-level metrics for a fair evaluation of texture generation. For semantic similarity, we use CLIP-based FID ($FID_{CLIP}$) via Clean-FID~\cite{parmar2022aliased}, and CLIP-MMD (CMMD)~\cite{jayasumana2024rethinking} for finer fidelity assessment. CLIP-Score~\cite{taited2023CLIPScore} measures alignment with input prompts, while LPIPS~\cite{zhang2018unreasonable} evaluates perceptual similarity to ground truth. We further include a human evaluation metric (“User-perf.”) to assess the perceptual quality of generated textures.

For tasks with strictly ground-truth supervision, such as second-stage evaluation from gt views, we assess texture quality using PSNR measured on the predicted surface via sampled UV points (denoted as PSNR\textsubscript{uv}). To evaluate the perceptual quality of the rendered views, we additionally report PSNR, SSIM~\cite{wang2004image}, and LPIPS~\cite{zhang2018unreasonable}.

\subsection{Qualitative  Comparision}
\subsubsection{Texture Generation Comparision} 
\label{sec:texgen_qualitative}
We evaluate our method visually across a diverse set of meshes, including raw scans, artist-created models, and outputs from mainstream proprietary 3D generation pipelines such as Tripo, Rodin, and Hunyuan 2.5. We compare against state-of-the-art image-based texturing approaches, including Paint3D~\cite{zeng2024paint3d} and Hunyuan2.0-Paint~\cite{zhao2025hunyuan3d}. Additionally, we include comparisons with proprietary methods such as Rodin\footnote{\url{https://hyper3d.ai/omnicraft/texture?lang=zh}} and Meshy\footnote{\url{https://www.meshy.ai/workspace}} to further assess texturing quality.
As demonstrated in Fig.~\ref{fig:main_comparision}, our method consistently achieves better performance across all baselines and mesh sources. Our model effectively extracts information from shaded images to restore fine-grained texture details, as clearly observed in the mask and Buddha examples. Secondly. It also better respects structural geometry, accurately recovering the door frames and handles of the vehicle while maintaining the desired rusted style — a capability lacking in other methods.
Additionally, our textures exhibit richer details and improved visual coherence, particularly in the cases shown in the fourth and fifth columns.

\subsubsection{Refinement Stage Comparision} 
\label{sec:refinement_stage_comp}
In this section, we evaluate the effectiveness of the second stage of our overall texturing pipeline. Specifically, we use the generated images from the first stage as input and compare different methods for texture refinement. We benchmark our approach against state-of-the-art UV-based methods, including Paint3D~\cite{zeng2024paint3d} and TexGen~\cite{yu2024texgen}. As shown in Fig.~\ref{fig:2nd_stage_comp} (best viewed by zooming in), our method consistently gains better results. In the first column, automatic UV unwrapping results in numerous small, fragmented regions on the face. UV-based methods struggle to refine these areas, leading to noticeable color inconsistencies and poor results. In contrast, our approach produces smooth and coherent textures. Furthermore, our model better respects the semantic consistency of texture sets. For example, in the first case, the glasses are seamlessly completed, and in the second case, both the ribs and the small emblem are accurately and coherently inpainted by our method.

\subsubsection{Stylized Texturing}
In this section, we demonstrate that our method can be applied to stylized texturing for 3D assets. Stylized texturing refers to the process of generating texture maps that not only align with the geometry of the 3D object but also reflect the visual style of a given reference image—such as oil painting, metallic finish, or cartoon shading. As illustrated in Fig.~\ref{fig_only_style}, our approach enables the creation of realistic and coherent textures while faithfully adapting to the appearance characteristics of different style exemplars.
\begin{table*}[t]
\centering
\tiny
\label{tab:lora_training_exp}
\resizebox{\linewidth}{!}{%
\begin{tabular}{lcccccccccc}
\toprule
\multirow{2}{*}{Methods.} & \multicolumn{4}{c}{Normal Estimation} & \multicolumn{6}{c}{MV Texture Generation} \\
\cmidrule(lr){2-5}\cmidrule(lr){6-11}
& mean$\downarrow$  & RMSE $\downarrow$& $\text{11.25}^\circ\uparrow$ & $\text{22.5}^\circ\uparrow$ & CMMD$\downarrow$ & FID$_{\text{CLIP}}$$\downarrow$ & CLIP$_{\text{score}}$$\uparrow$ & LPIPS$\downarrow$ & Mem. Cost$\downarrow$ & Speed$\downarrow$\\
\midrule
w/o Drop Training   & \textbf{22.57} & \textbf{29.42} & 44.85 & \textbf{55.82} & 0.912&17.49& 0.839&\textbf{0.087} & 69.2GB &38.76s/it\\
w/ Drop Training  & 22.79& 29.79& \textbf{45.13}&55.48 & \textbf{0.826} & \textbf{16.03}  &\textbf{0.844} & 0.090 &  \textbf{53.6GB} &\textbf{21.50s/it} \\
\bottomrule
\end{tabular}%
}
\caption{Ablation studies of MV texture generation and normal estimation using drop training strategy.}
    \vspace{-0.5cm}
\label{tab:lora_training_comp}
\end{table*}
\subsection{Quantitative Comparision}
\subsubsection{Texture Generation Comparision}
As shown in Tab.~\ref{tab:texture_main_comparison}, our method outperforms the current state-of-the-art across both benchmarks. While Paint3D achieves a comparable CLIP score on artist-created meshes, its performance degrades significantly on generative models, highlighting the limited generalizability of UV-based mapping methods discussed in the introduction. In contrast, our approach consistently yields stronger results among two-stage pipelines. Additionally, TexGaussians performs relatively well in real track but poorly on in-the-wild generative data, likely due to the limited availability of large scale high-quality native 3D texture datasets. These findings indicate that native 3D approaches are currently better suited for refinement or inpainting, rather than as standalone generative solutions. Therefore, leveraging 2D diffusion priors remains critical. Our method achieves highly competitive results across diverse methods, demonstrating its robustness in texturing 3D shapes completely with varying topologies.

\begin{table}[t]
\centering
\resizebox{0.9\linewidth}{!}{%
\begin{tabular}{c|cc}
\toprule
Methods.& PSNR\textsubscript{uv}$\uparrow$&PSNR$\uparrow$\\
\midrule
 w/o Texture Function Supervision & 20.31  &  25.81 \\
 w Texture Function Supervision &\textbf{20.99}& \textbf{27.01}\\
\bottomrule
\end{tabular}%
}
\caption{Ablation studies on Texture function.}
\vspace{-0.5cm}
\label{tab:ltm_ablation}
\end{table}

\subsubsection{Refinement Stage Comparision} 
We also conduct a quantitative evaluation of our refinement stage on the artist-created mesh benchmark to rigorously assess the completeness and quality of the texture produced by our pipeline. As shown in Tab.~\ref{tab:stage2comp}, the results demonstrate that our second stage achieves the best performance in both preserving visible regions and completing occluded areas. In addition, our method demonstrates strong performance in terms of rendered image quality, further validating the visual fidelity of the generated textures. These results suggest that our approach provides a more effective refinement paradigm compared to UV-space operations or methods that treat texture purely as a 2D image. This is because texture appearance is often closely tied to the underlying geometry, and operating in a 3D functional space allows our method to better incorporate geometric structure during refinement. Overall, the findings further support our claim that operate texture in 3D functional space is more suitable to represent texture in UV-space.

\subsection{Ablation Study} %
In this section, we systematically evaluate all our proposed key components, which includes a comparison of the effectiveness of the patch dropout strategy (Sec.\ref{sec:diffusion_training_strategy}), and the impact of texture function supervision (Sec.\ref{sec:tf_ablation}). It is important to note that, due to limitations in computational resources, certain ablation experiments were performed using fewer training iterations than the full model. Further implementation details and discussion are in our Appendix.
\begin{figure}[!t]
    \centering
  \begin{minipage}[b]{\linewidth}
    \centering
    \includegraphics[width=\linewidth]{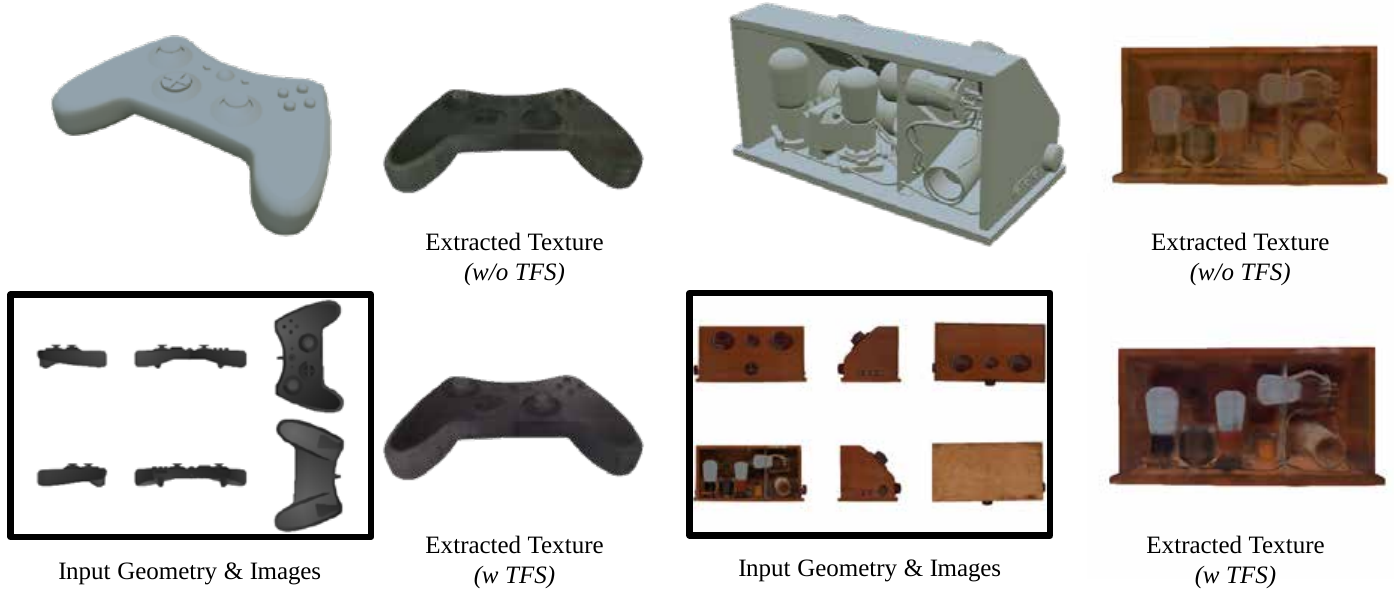}
  \end{minipage}
      \vspace{-0.5cm}
    \caption{ Visualization of the effectiveness of Texture Function Supervision (TFS). Under identical training iterations, models trained with TFS yield significantly higher-quality and completed textures compared to those supervised solely on surface signals. (Best viewed when zoomed in).}
    \label{fig:vis-tfablation}
    \vspace{-0.5cm}
\end{figure}
\subsubsection{Drop Training Strategy}
\label{sec:diffusion_training_strategy}
In this section, we evaluate our proposed drop training strategy. 
We first evaluate our drop training strategy on texture generation task. In addition, we introduce the normal estimation task as a complementary evaluation to investigate advanced training strategies for large-scale diffusion transformers.
To assess the impact of our strategy, we conduct a controlled experiment. We train the same model on the same dataset for an identical number of iterations (e.g., 1500), with the only variable being the presence or absence of our drop training strategy.

For texture generation, we adopt the Artist-Created Mesh comparison setting described in Sec. \ref{sec:tex_main_experiments}. For normal estimation, we follow the evaluation protocol used by GeoWizard~\cite{ke2024repurposing}, which employs metrics as shown in Tab.~\ref{tab:lora_training_comp}. Further details on the settings can be found in the Appendix. As shown in Tab.~\ref{tab:lora_training_comp}, our experiments indicate that using dropout during training can produce comparable performance in 3 tasks we select, while accelerating training by a significant percentage specifically, drop 50$\%$ tokens in MV texture generation task can save 22.5$\%$ memory cost and speed up our training about 44.5$\%$, (evaluate using A800 bs=4). These experimental results in various tasks indicate that our method is a useful plug-and-play training strategy for fine-tuning diffusion models.

\subsubsection{Texture Functions Supervision v.s. Surface Supervision}
\label{sec:tf_ablation}
In this section, we evaluate the effectiveness of using texture functions (TFs) as supervision within our LTM framework. A straightforward alternative is to directly supervise the LTM using rendered images or surface-sampled RGB values, as adopted in prior works such as~\cite{oechsle2019texture, jiang2025dimer, yu2023texture}. For TF-based supervision, we utilize queries from the completed 3D volume, leveraging its well-defined volumetric representation. In contrast, the baseline approach limits supervision to surface points only. It is worth noting that, unlike the setup in Tab.~\ref{tab:stage2comp}, this evaluation textures the entire model directly through LTM query points, without blending with partial texture geometry. This allows for a comprehensive assessment of the underlying representation's effectiveness. 

As shown in Tab.~\ref{tab:ltm_ablation} and Fig.~\ref{fig:vis-tfablation}, supervising with texture functions consistently yields superior performance, achieving the highest PSNR and lowest LPIPS, which reflects improved fidelity and perceptual quality. In particular, the highest PSNR\textsubscript{uv} scores indicate a significant enhancement in texture completeness. These results demonstrate the effectiveness of our proposed TFs.

\section{Conclusion}
We presented UniTEX, a universal and general framework to generate high fidelity texture for 3D shapes. Unlike previous UV-inpainting based methods, which are limited by inherent problems like topological ambiguity in UV mapping, we proposed to bypass UV unwarping entirely. This is achieved through our Texture Functions (TFs), a topology-agnostic and continuous 3D texture representation that models texture as a complete spatial field. We then designed a Large Texturing Model (LTM) to effectively predict these TFs directly from geometries, multi-view inputs and partial textured models, enabling robust and high-fidelity texture completion. Furthermore, we developed an efficient LoRA-based strategy to adapt large-scale Diffusion Transformers (DiTs) for advanced multi-view texture synthesis. Extensive experiments on benchmarks including both artist-created and challenging generative 3D models demonstrate UniTEX’s superior consistency, perceptual quality, and scalability compared to prior 3D texturing methods.
{
    \small
    \bibliographystyle{ieeenat_fullname}
    \bibliography{main}

\begin{thebibliography}{48}
\providecommand{\natexlab}[1]{#1}
\providecommand{\url}[1]{\texttt{#1}}
\expandafter\ifx\csname urlstyle\endcsname\relax
  \providecommand{\doi}[1]{doi: #1}\else
  \providecommand{\doi}{doi: \begingroup \urlstyle{rm}\Url}\fi

\bibitem[Bensadoun et~al.(2024)Bensadoun, Kleiman, Azuri, Harosh, Vedaldi, Neverova, and Gafni]{bensadoun2024metatexgen}
Raphael Bensadoun, Yanir Kleiman, Idan Azuri, Omri Harosh, Andrea Vedaldi, Natalia Neverova, and Oran Gafni.
\newblock Meta 3d texturegen: Fast and consistent texture generation for 3d objects.
\newblock \emph{arXiv preprint arXiv:2407.02430}, 2024.

\bibitem[Cao et~al.(2023)Cao, Kreis, Fidler, Sharp, and Yin]{cao2023texfusion}
Tianshi Cao, Karsten Kreis, Sanja Fidler, Nicholas Sharp, and Kangxue Yin.
\newblock Texfusion: Synthesizing 3d textures with text-guided image diffusion models.
\newblock In \emph{Proceedings of the IEEE/CVF International Conference on Computer Vision}, pages 4169--4181, 2023.

\bibitem[Chen et~al.(2023{\natexlab{a}})Chen, Siddiqui, Lee, Tulyakov, and Nie{\ss}ner]{chen2023text2tex}
Dave~Zhenyu Chen, Yawar Siddiqui, Hsin-Ying Lee, Sergey Tulyakov, and Matthias Nie{\ss}ner.
\newblock Text2tex: Text-driven texture synthesis via diffusion models.
\newblock In \emph{Proceedings of the IEEE/CVF international conference on computer vision}, pages 18558--18568, 2023{\natexlab{a}}.

\bibitem[Chen et~al.(2023{\natexlab{b}})Chen, Chen, Jiao, and Jia]{chen2023fantasia3d}
Rui Chen, Yongwei Chen, Ningxin Jiao, and Kui Jia.
\newblock Fantasia3d: Disentangling geometry and appearance for high-quality text-to-3d content creation.
\newblock In \emph{ICCV}, 2023{\natexlab{b}}.

\bibitem[Chen et~al.(2024)Chen, Zhang, Liang, Luo, Li, Liu, Li, Long, Feng, and Tan]{chen2024dora}
Rui Chen, Jianfeng Zhang, Yixun Liang, Guan Luo, Weiyu Li, Jiarui Liu, Xiu Li, Xiaoxiao Long, Jiashi Feng, and Ping Tan.
\newblock Dora: Sampling and benchmarking for 3d shape variational auto-encoders.
\newblock \emph{arXiv preprint arXiv:2412.17808}, 2024.

\bibitem[Chen et~al.(2023{\natexlab{c}})Chen, Qian, Tang, Lai, Liu, Han, and Jia]{chen2023longlora}
Yukang Chen, Shengju Qian, Haotian Tang, Xin Lai, Zhijian Liu, Song Han, and Jiaya Jia.
\newblock Longlora: Efficient fine-tuning of long-context large language models.
\newblock \emph{arXiv preprint arXiv:2309.12307}, 2023{\natexlab{c}}.

\bibitem[Cheng et~al.(2024)Cheng, Mu, Zeng, Chen, Pang, Zhang, Wang, Fu, Yu, Liu, and Pan]{cheng2024mvpaint}
Wei Cheng, Juncheng Mu, Xianfang Zeng, Xin Chen, Anqi Pang, Chi Zhang, Zhibin Wang, Bin Fu, Gang Yu, Ziwei Liu, and Liang Pan.
\newblock Mvpaint: Synchronized multi-view diffusion for painting anything 3d.
\newblock \emph{arXiv preprint arxiv:2411.02336}, 2024.

\bibitem[Deitke et~al.(2023)Deitke, Schwenk, Salvador, Weihs, Michel, VanderBilt, Schmidt, Ehsani, Kembhavi, and Farhadi]{deitke2023objaverse}
Matt Deitke, Dustin Schwenk, Jordi Salvador, Luca Weihs, Oscar Michel, Eli VanderBilt, Ludwig Schmidt, Kiana Ehsani, Aniruddha Kembhavi, and Ali Farhadi.
\newblock Objaverse: A universe of annotated 3d objects.
\newblock In \emph{Proceedings of the IEEE/CVF Conference on Computer Vision and Pattern Recognition}, pages 13142--13153, 2023.

\bibitem[Deng et~al.(2024)Deng, Omernick, Weiss, Ramanan, Zhu, Zhou, and Agrawala]{deng2024flashtex}
Kangle Deng, Timothy Omernick, Alexander Weiss, Deva Ramanan, Jun-Yan Zhu, Tinghui Zhou, and Maneesh Agrawala.
\newblock Flashtex: Fast relightable mesh texturing with lightcontrolnet.
\newblock In \emph{European Conference on Computer Vision}, pages 90--107. Springer, 2024.

\bibitem[Dong et~al.(2024)Dong, Fan, Guo, Wang, Zhang, Chen, Luo, and Zou]{dong2024tsd}
Linwei Dong, Qingnan Fan, Yihong Guo, Zhonghao Wang, Qi Zhang, Jinwei Chen, Yawei Luo, and Changqing Zou.
\newblock Tsd-sr: One-step diffusion with target score distillation for real-world image super-resolution.
\newblock \emph{arXiv preprint arXiv:2411.18263}, 2024.

\bibitem[Esser et~al.(2024)Esser, Kulal, Blattmann, Entezari, M{\"u}ller, Saini, Levi, Lorenz, Sauer, Boesel, et~al.]{esser2024sd3}
Patrick Esser, Sumith Kulal, Andreas Blattmann, Rahim Entezari, Jonas M{\"u}ller, Harry Saini, Yam Levi, Dominik Lorenz, Axel Sauer, Frederic Boesel, et~al.
\newblock Scaling rectified flow transformers for high-resolution image synthesis.
\newblock In \emph{Forty-first international conference on machine learning}, 2024.

\bibitem[Guo et~al.(2025)Guo, Gao, Bian, Sun, Zheng, Jia, and Gong]{guo2025hyper3d}
Jingyu Guo, Sensen Gao, Jia-Wang Bian, Wanhu Sun, Heliang Zheng, Rongfei Jia, and Mingming Gong.
\newblock Hyper3d: Efficient 3d representation via hybrid triplane and octree feature for enhanced 3d shape variational auto-encoders.
\newblock \emph{arXiv preprint arXiv:2503.10403}, 2025.

\bibitem[Hong et~al.(2023)Hong, Zhang, Gu, Bi, Zhou, Liu, Liu, Sunkavalli, Bui, and Tan]{hong2023lrm}
Yicong Hong, Kai Zhang, Jiuxiang Gu, Sai Bi, Yang Zhou, Difan Liu, Feng Liu, Kalyan Sunkavalli, Trung Bui, and Hao Tan.
\newblock Lrm: Large reconstruction model for single image to 3d.
\newblock \emph{arXiv preprint arXiv:2311.04400}, 2023.

\bibitem[Jayasumana et~al.(2024)Jayasumana, Ramalingam, Veit, Glasner, Chakrabarti, and Kumar]{jayasumana2024rethinking}
Sadeep Jayasumana, Srikumar Ramalingam, Andreas Veit, Daniel Glasner, Ayan Chakrabarti, and Sanjiv Kumar.
\newblock Rethinking fid: Towards a better evaluation metric for image generation.
\newblock In \emph{Proceedings of the IEEE/CVF Conference on Computer Vision and Pattern Recognition}, pages 9307--9315, 2024.

\bibitem[Jiang et~al.(2025)Jiang, Lin, Chen, Ge, Yang, Jiang, Lyu, Zheng, and Chen]{jiang2025dimer}
Lutao Jiang, Jiantao Lin, Kanghao Chen, Wenhang Ge, Xin Yang, Yifan Jiang, Yuanhuiyi Lyu, Xu Zheng, and Yingcong Chen.
\newblock Dimer: Disentangled mesh reconstruction model.
\newblock \emph{arXiv preprint arXiv:2504.17670}, 2025.

\bibitem[Ke et~al.(2024)Ke, Obukhov, Huang, Metzger, Daudt, and Schindler]{ke2024repurposing}
Bingxin Ke, Anton Obukhov, Shengyu Huang, Nando Metzger, Rodrigo~Caye Daudt, and Konrad Schindler.
\newblock Repurposing diffusion-based image generators for monocular depth estimation.
\newblock In \emph{Proceedings of the IEEE/CVF Conference on Computer Vision and Pattern Recognition}, pages 9492--9502, 2024.

\bibitem[Li et~al.(2023{\natexlab{a}})Li, Li, Zhu, Yu, Zhao, Wan, You, Shi, and Lin]{li2023instant3d}
Sixu Li, Chaojian Li, Wenbo Zhu, Boyang Yu, Yang Zhao, Cheng Wan, Haoran You, Huihong Shi, and Yingyan Lin.
\newblock Instant-3d: Instant neural radiance field training towards on-device ar/vr 3d reconstruction.
\newblock In \emph{Proceedings of the 50th Annual International Symposium on Computer Architecture}, pages 1--13, 2023{\natexlab{a}}.

\bibitem[Li et~al.(2023{\natexlab{b}})Li, Chen, Chen, and Tan]{li2023sweetdreamer}
Weiyu Li, Rui Chen, Xuelin Chen, and Ping Tan.
\newblock Sweetdreamer: Aligning geometric priors in 2d diffusion for consistent text-to-3d.
\newblock \emph{arXiv preprint arXiv:2310.02596}, 2023{\natexlab{b}}.

\bibitem[Li et~al.(2024)Li, Liu, Chen, Liang, Chen, Tan, and Long]{li2024craftsman}
Weiyu Li, Jiarui Liu, Rui Chen, Yixun Liang, Xuelin Chen, Ping Tan, and Xiaoxiao Long.
\newblock Craftsman: High-fidelity mesh generation with 3d native generation and interactive geometry refiner.
\newblock \emph{arXiv preprint arXiv:2405.14979}, 2024.

\bibitem[Li et~al.(2025)Li, Zou, Liu, Wang, Liang, Yu, Liu, Guo, Liang, Ouyang, et~al.]{li2025triposg}
Yangguang Li, Zi-Xin Zou, Zexiang Liu, Dehu Wang, Yuan Liang, Zhipeng Yu, Xingchao Liu, Yuan-Chen Guo, Ding Liang, Wanli Ouyang, et~al.
\newblock Triposg: High-fidelity 3d shape synthesis using large-scale rectified flow models.
\newblock \emph{arXiv preprint arXiv:2502.06608}, 2025.

\bibitem[Lin et~al.(2023)Lin, Gao, Tang, Takikawa, Zeng, Huang, Kreis, Fidler, Liu, and Lin]{lin2023magic3d}
Chen-Hsuan Lin, Jun Gao, Luming Tang, Towaki Takikawa, Xiaohui Zeng, Xun Huang, Karsten Kreis, Sanja Fidler, Ming-Yu Liu, and Tsung-Yi Lin.
\newblock Magic3d: High-resolution text-to-3d content creation.
\newblock In \emph{CVPR}, 2023.

\bibitem[Liu et~al.(2024{\natexlab{a}})Liu, Wu, Liu, Liu, Wu, Peng, Zhao, Feng, Liu, and Ding]{Liu2024TexOct}
Jialun Liu, Chenming Wu, Xinqi Liu, Xing Liu, Jinbo Wu, Haotian Peng, Chen Zhao, Haocheng Feng, Jingtuo Liu, and Errui Ding.
\newblock Texoct: Generating textures of 3d models with octree-based diffusion.
\newblock In \emph{Proceedings of the IEEE/CVF Conference on Computer Vision and Pattern Recognition (CVPR)}, pages 4284--4293, 2024{\natexlab{a}}.

\bibitem[Liu et~al.(2023)Liu, Shi, Chen, Zhang, Xu, Wei, Chen, Zeng, Gu, and Su]{liu2023one12345++}
Minghua Liu, Ruoxi Shi, Linghao Chen, Zhuoyang Zhang, Chao Xu, Xinyue Wei, Hansheng Chen, Chong Zeng, Jiayuan Gu, and Hao Su.
\newblock One-2-3-45++: Fast single image to 3d objects with consistent multi-view generation and 3d diffusion.
\newblock \emph{arXiv preprint arXiv:2311.07885}, 2023.

\bibitem[Liu et~al.(2024{\natexlab{b}})Liu, Zeng, Wei, Shi, Chen, Xu, Zhang, Wang, Zhang, Liu, et~al.]{liu2024meshformer}
Minghua Liu, Chong Zeng, Xinyue Wei, Ruoxi Shi, Linghao Chen, Chao Xu, Mengqi Zhang, Zhaoning Wang, Xiaoshuai Zhang, Isabella Liu, et~al.
\newblock Meshformer: High-quality mesh generation with 3d-guided reconstruction model.
\newblock \emph{arXiv preprint arXiv:2408.10198}, 2024{\natexlab{b}}.

\bibitem[Liu et~al.(2024{\natexlab{c}})Liu, Xie, Liu, and Wong]{liu2024syncmvd}
Yuxin Liu, Minshan Xie, Hanyuan Liu, and Tien-Tsin Wong.
\newblock Text-guided texturing by synchronized multi-view diffusion.
\newblock In \emph{SIGGRAPH Asia 2024 Conference Papers}, pages 1--11, 2024{\natexlab{c}}.

\bibitem[Lorensen and Cline(1998)]{lorensen1998marching}
William~E Lorensen and Harvey~E Cline.
\newblock Marching cubes: A high resolution 3d surface construction algorithm.
\newblock In \emph{Seminal graphics: pioneering efforts that shaped the field}, pages 347--353. 1998.

\bibitem[Oechsle et~al.(2019)Oechsle, Mescheder, Niemeyer, Strauss, and Geiger]{oechsle2019texture}
Michael Oechsle, Lars Mescheder, Michael Niemeyer, Thilo Strauss, and Andreas Geiger.
\newblock Texture fields: Learning texture representations in function space.
\newblock In \emph{Proceedings of the IEEE/CVF international conference on computer vision}, pages 4531--4540, 2019.

\bibitem[Parmar et~al.(2022)Parmar, Zhang, and Zhu]{parmar2022aliased}
Gaurav Parmar, Richard Zhang, and Jun-Yan Zhu.
\newblock On aliased resizing and surprising subtleties in gan evaluation.
\newblock In \emph{Proceedings of the IEEE/CVF conference on computer vision and pattern recognition}, pages 11410--11420, 2022.

\bibitem[Reiser et~al.(2023)Reiser, Szeliski, Verbin, Srinivasan, Mildenhall, Geiger, Barron, and Hedman]{reiser2023merf}
Christian Reiser, Rick Szeliski, Dor Verbin, Pratul Srinivasan, Ben Mildenhall, Andreas Geiger, Jon Barron, and Peter Hedman.
\newblock Merf: Memory-efficient radiance fields for real-time view synthesis in unbounded scenes.
\newblock \emph{ACM Transactions on Graphics (TOG)}, 42\penalty0 (4):\penalty0 1--12, 2023.

\bibitem[Richardson et~al.(2023)Richardson, Metzer, Alaluf, Giryes, and Cohen-Or]{richardson2023texture}
Elad Richardson, Gal Metzer, Yuval Alaluf, Raja Giryes, and Daniel Cohen-Or.
\newblock Texture: Text-guided texturing of 3d shapes.
\newblock In \emph{ACM SIGGRAPH 2023 conference proceedings}, pages 1--11, 2023.

\bibitem[Rombach et~al.(2022)Rombach, Blattmann, Lorenz, Esser, and Ommer]{rombach2022high}
Robin Rombach, Andreas Blattmann, Dominik Lorenz, Patrick Esser, and Bj{\"o}rn Ommer.
\newblock High-resolution image synthesis with latent diffusion models.
\newblock In \emph{CVPR}, 2022.

\bibitem[Siddiqui et~al.(2022)Siddiqui, Thies, Ma, Shan, Nie{\ss}ner, and Dai]{siddiqui2022texturify}
Yawar Siddiqui, Justus Thies, Fangchang Ma, Qi Shan, Matthias Nie{\ss}ner, and Angela Dai.
\newblock Texturify: Generating textures on 3d shape surfaces.
\newblock In \emph{European Conference on Computer Vision}, pages 72--88. Springer, 2022.

\bibitem[Tan et~al.(2024)Tan, Liu, Yang, Xue, and Wang]{tan2024ominicontrol}
Zhenxiong Tan, Songhua Liu, Xingyi Yang, Qiaochu Xue, and Xinchao Wang.
\newblock Ominicontrol: Minimal and universal control for diffusion transformer.
\newblock \emph{arXiv preprint arXiv:2411.15098}, 2024.

\bibitem[Tang et~al.(2024)Tang, Chen, Wang, Tang, Zhang, Fan, Chandra, Furukawa, and Ranjan]{tang2024mvdiffusion++}
Shitao Tang, Jiacheng Chen, Dilin Wang, Chengzhou Tang, Fuyang Zhang, Yuchen Fan, Vikas Chandra, Yasutaka Furukawa, and Rakesh Ranjan.
\newblock Mvdiffusion++: A dense high-resolution multi-view diffusion model for single or sparse-view 3d object reconstruction.
\newblock In \emph{European Conference on Computer Vision}, pages 175--191. Springer, 2024.

\bibitem[Wang et~al.(2004)Wang, Bovik, Sheikh, and Simoncelli]{wang2004image}
Zhou Wang, Alan~C Bovik, Hamid~R Sheikh, and Eero~P Simoncelli.
\newblock Image quality assessment: from error visibility to structural similarity.
\newblock \emph{TIP}, 2004.

\bibitem[Wang et~al.(2024)Wang, Wang, Chen, Xiang, Chen, Yu, Li, Su, and Zhu]{wang2024crm}
Zhengyi Wang, Yikai Wang, Yifei Chen, Chendong Xiang, Shuo Chen, Dajiang Yu, Chongxuan Li, Hang Su, and Jun Zhu.
\newblock Crm: Single image to 3d textured mesh with convolutional reconstruction model.
\newblock \emph{arXiv preprint arXiv:2403.05034}, 2024.

\bibitem[Xiong et~al.(2024)Xiong, Liu, Hu, Wu, Wu, Liu, Zhao, Ding, and Lian]{xiong2024texgaussian}
Bojun Xiong, Jialun Liu, Jiakui Hu, Chenming Wu, Jinbo Wu, Xing Liu, Chen Zhao, Errui Ding, and Zhouhui Lian.
\newblock Texgaussian: Generating high-quality pbr material via octree-based 3d gaussian splatting.
\newblock \emph{arXiv preprint arXiv:2411.19654}, 2024.

\bibitem[Yu et~al.(2023)Yu, Dai, Li, Ma, Liu, and Qi]{yu2023texture}
Xin Yu, Peng Dai, Wenbo Li, Lan Ma, Zhengzhe Liu, and Xiaojuan Qi.
\newblock Texture generation on 3d meshes with point-uv diffusion.
\newblock In \emph{Proceedings of the IEEE/CVF International Conference on Computer Vision}, pages 4206--4216, 2023.

\bibitem[Yu et~al.(2024)Yu, Yuan, Guo, Liu, Liu, Li, Cao, Liang, and Qi]{yu2024texgen}
Xin Yu, Ze Yuan, Yuan-Chen Guo, Ying-Tian Liu, Jianhui Liu, Yangguang Li, Yan-Pei Cao, Ding Liang, and Xiaojuan Qi.
\newblock Texgen: a generative diffusion model for mesh textures.
\newblock \emph{ACM Transactions on Graphics (TOG)}, 43\penalty0 (6):\penalty0 1--14, 2024.

\bibitem[Zeng et~al.(2024)Zeng, Chen, Qi, Liu, Zhao, Wang, Fu, Liu, and Yu]{zeng2024paint3d}
Xianfang Zeng, Xin Chen, Zhongqi Qi, Wen Liu, Zibo Zhao, Zhibin Wang, Bin Fu, Yong Liu, and Gang Yu.
\newblock Paint3d: Paint anything 3d with lighting-less texture diffusion models.
\newblock In \emph{Proceedings of the IEEE/CVF conference on computer vision and pattern recognition}, pages 4252--4262, 2024.

\bibitem[Zhang et~al.(2023)Zhang, Tang, Niessner, and Wonka]{zhang20233dshape2vecset}
Biao Zhang, Jiapeng Tang, Matthias Niessner, and Peter Wonka.
\newblock 3dshape2vecset: A 3d shape representation for neural fields and generative diffusion models.
\newblock \emph{ACM Transactions on Graphics (TOG)}, 42\penalty0 (4):\penalty0 1--16, 2023.

\bibitem[Zhang et~al.(2024{\natexlab{a}})Zhang, Pan, Zhang, Zhu, and Gao]{zhang2024texpainter}
Hongkun Zhang, Zherong Pan, Congyi Zhang, Lifeng Zhu, and Xifeng Gao.
\newblock Texpainter: Generative mesh texturing with multi-view consistency.
\newblock In \emph{ACM SIGGRAPH 2024 Conference Papers}, pages 1--11, 2024{\natexlab{a}}.

\bibitem[Zhang et~al.(2024{\natexlab{b}})Zhang, Wang, Zhang, Qiu, Pang, Jiang, Yang, Xu, and Yu]{zhang2024clay}
Longwen Zhang, Ziyu Wang, Qixuan Zhang, Qiwei Qiu, Anqi Pang, Haoran Jiang, Wei Yang, Lan Xu, and Jingyi Yu.
\newblock Clay: A controllable large-scale generative model for creating high-quality 3d assets.
\newblock \emph{ACM Transactions on Graphics (TOG)}, 43\penalty0 (4):\penalty0 1--20, 2024{\natexlab{b}}.

\bibitem[Zhang et~al.(2018)Zhang, Isola, Efros, Shechtman, and Wang]{zhang2018unreasonable}
Richard Zhang, Phillip Isola, Alexei~A Efros, Eli Shechtman, and Oliver Wang.
\newblock The unreasonable effectiveness of deep features as a perceptual metric.
\newblock In \emph{Proceedings of the IEEE conference on computer vision and pattern recognition}, pages 586--595, 2018.

\bibitem[Zhang et~al.(2024{\natexlab{c}})Zhang, Liu, Xie, Yang, Liu, Yang, Zhang, Kou, Lin, Wang, et~al.]{zhang2024dreammat}
Yuqing Zhang, Yuan Liu, Zhiyu Xie, Lei Yang, Zhongyuan Liu, Mengzhou Yang, Runze Zhang, Qilong Kou, Cheng Lin, Wenping Wang, et~al.
\newblock Dreammat: High-quality pbr material generation with geometry-and light-aware diffusion models.
\newblock \emph{ACM Transactions on Graphics (TOG)}, 43\penalty0 (4):\penalty0 1--18, 2024{\natexlab{c}}.

\bibitem[Zhang et~al.(2025)Zhang, Yuan, Song, Wang, and Liu]{zhang2025easycontrol}
Yuxuan Zhang, Yirui Yuan, Yiren Song, Haofan Wang, and Jiaming Liu.
\newblock Easycontrol: Adding efficient and flexible control for diffusion transformer.
\newblock \emph{arXiv preprint arXiv:2503.07027}, 2025.

\bibitem[Zhao et~al.(2025)Zhao, Lai, Lin, Zhao, Liu, Yang, Feng, Yang, Zhang, Yang, et~al.]{zhao2025hunyuan3d}
Zibo Zhao, Zeqiang Lai, Qingxiang Lin, Yunfei Zhao, Haolin Liu, Shuhui Yang, Yifei Feng, Mingxin Yang, Sheng Zhang, Xianghui Yang, et~al.
\newblock Hunyuan3d 2.0: Scaling diffusion models for high resolution textured 3d assets generation.
\newblock \emph{arXiv preprint arXiv:2501.12202}, 2025.

\bibitem[Zhengwentai(2023)]{taited2023CLIPScore}
SUN Zhengwentai.
\newblock {clip-score: CLIP Score for PyTorch}.
\newblock \url{https://github.com/taited/clip-score}, 2023.
\newblock Version 0.2.1.

\end{thebibliography}
}

\end{document}